\documentclass[a4paper,fleqn]{cas-dc}

\usepackage[numbers]{natbib}

\def\tsc#1{\csdef{#1}{\textsc{\lowercase{#1}}\xspace}}
\tsc{WGM}
\tsc{QE}

\begin{document}
\let\WriteBookmarks\relax
\def\floatpagepagefraction{1}
\def\textpagefraction{.001}

\shorttitle{}    

\shortauthors{Ren-Dong Xie et~al.}  

\title [mode = title]{Semantic-Aware Representation Learning via Conditional Transport for Multi-Label Image Classification}  



\author[1,2]{Ren-Dong Xie}[orcid=0009-0003-3059-1657]
\ead{silencexie69@163.com}
\credit{Supervision, Conceptualization, Methodology, Software.}

\author[1,2]{Zhi-Fen He}[orcid=0000-0003-1969-1772]
\cormark[1]
\ead{zfhe323@163.com}
\credit{Data curation, Writing – original draft, Writing – review & editing.}

\author[1,2]{Bo Li}[orcid=0000-0001-9025-8985]
\ead{libo@nchu.edu.cn}
\credit{Conceptualization, Methodology, Software, Visualization.}

\author[1,2]{Bin Liu}[orcid=0000-0002-5023-7167]
\ead{nyliubin@nchu.edu.cn}
\credit{Writing – review & editing, Supervision.}

\author[1,2]{Jin-Yan Hu}[orcid=0009-0003-4284-5220]
\ead{jyh1472580628@163.com}
\credit{Software, Validation.}

\affiliation[1]{organization={School of Mathematics and Information Science},
    addressline={Nanchang Hangkong University}, 
    city={Nanchang},
    postcode={330063}, 
    country={China}}
\affiliation[2]{organization={Key Laboratory of Jiangxi Province for Image Processing and Pattern Recognition},
    addressline={Nanchang Hangkong University}, 
    city={Nanchang},
    postcode={330063}, 
    country={China}}

\cortext[cor1]{Corresponding author: Tel.: +86 0791-83863755; Postal address: No.696, Fenghe South Road, Honggutan District, Nanchang, Jiangxi
330063, China.}



\begin{abstract}
Multi-label image classification is a critical task in machine learning that aims to accurately assign multiple labels to a single image. While existing methods often utilize attention mechanisms or graph convolutional networks to model visual representations, their performance is still constrained by two critical limitations: the inability to learn discriminative semantic-aware features, and the lack of fine-grained alignment between visual representations and label embeddings. To tackle these issues in a unified framework, this paper proposes a novel approach named \textbf{S}emantic-aware representation learning via \textbf{C}onditional \textbf{T}ransport for Multi-Label Image Classification (SCT). The proposed method introduces a semantic-related feature learning module that extracts discriminative label-specific features by emphasizing semantic relevance and interaction, along with a conditional transport-based alignment mechanism that enables precise visual-semantic alignment. Extensive experiments on two widely-used benchmark datasets, VOC2007 and MS-COCO, validate the effectiveness of SCT and demonstrate its superior performance compared to existing state-of-the-art methods.
\end{abstract}





\maketitle

\section{Introduction}\label{sec1}

In recent years, multi-label image classification (MLIC) has garnered significant research interest. Its goal is to determine all relevant semantic objects or concepts within a single image, which is widely applied in domains such as medical image analysis\cite{re1}, image retrieval\cite{re2}, and scene understanding\cite{re3}. 



Some methods learn visual features directly through deep neural networks for multi-label classification \cite{re4,re5,re6,re7}. For example, Guo et al.\cite{re4} leverage class activation maps (CAMs)\cite{re5} combined with an attention consistency loss\cite{re6} to impose invariance constraints on the network, thereby producing more robust visual features. Gao et al.\cite{re7} also utilize CAMs to identify and crop local regions corresponding to semantic objects. These regions are then up-sampled and fed into a weight‐sharing feature extractor to enhance object-relevant features. Although these approaches build stronger visual features, they work without semantic guidance, limiting their ability to distinguish different classes effectively.

Some methods focus on semantic alignment by treating image patches as tokens in Transformers, where label and patch embeddings interact via self-attention \cite{re14,re30}. For instance, Lanchantin et al. \cite{re14} explore label-image interactions via attention mechanism. Alexey et al. \cite{re30} adopt multi-head self-attention to learn global visual representations. However, these need complex setups and precise alignment. Li et al. \cite{re34} employ conditional transport to achieve more efficient feature alignment. However, their approach fails to learn label-specific features, which ultimately constrains its overall performance.

In this paper, we propose SCT, a \textbf{S}emantic-aware representation learning method via \textbf{C}onditional \textbf{T}ransport for Multi-Label Image Classification. First, image features are fused with label embeddings via global pooling to obtain semantic-related features. Then, conditional transport is leveraged to aligns these with initial features under semantic map guidance, producing refined visual representations. Finally, multi-label predictions are generated by aggregating region-level scores.


The main contributions of this paper are summarized as follows:

1) We propose SCT, a novel semantic‐aware representation learning framework that leverages conditional transport to  enhance multi‐label image classification by obtaining highly discriminative label-specific representations.

2) We design a conditional transport-based alignment mechanism that effectively models visual–semantic interaction, significantly improving the discriminability of the learned representations.

3) Extensive experiments on public benchmarks (VOC2007 and MS-COCO) demonstrate that SCT achieves superior performance over the state‐of‐the‐art multi-label classification methods.

\section{Related Work}\label{sec2}

\subsection{Label Correlation‐Based Methods}

Previous multi-label image classification methods typically trained separate binary classifiers for each label \cite{re15,re16,re17}, treating the task as multiple independent binary problems while ignoring label correlations. To address this limitation, some studies \cite{re18,re19} employed RNNs or LSTMs to jointly embed images and labels in a shared space and predict labels in a fixed sequence. However, these sequence-based approaches remain sensitive to label order and struggle to capture complex label relationships.Recent graph-based methods \cite{re8,re9,re10,re11,re20,re21,re22,re23,re24} have shown better capability in modeling label correlations. Chen et al. \cite{re8} fused label embeddings via bilinear attention \cite{re21} to identify salient regions, then refined features through graph propagation \cite{re9}. Subsequently, Chen et al. proposed two variants: C-GCN \cite{re10} and P-GCN \cite{re11} to model asymmetric label relationships. Deng et al.\cite{re22} utilized a multi‐headed GCN to learn classifiers from two heterogeneous sources of prior‐knowledge. Zhou et al.\cite{re23} proposed a Dual‐Relation Graph Network (DRGN), which consists of two complementary components: an Intra‐Image Spatial Exploration (ISE) module that  models label correlations via object‐level GCN, and a Cross‐Image Semantic Learning (CSL) module that refines features using image‐level GCN. You et al.\cite{re24} proposed an Adjacency‐based Semantic Graph Embedding (ASGE) to encode label semantics, combined with Cross‐Modal Attention (CMA) for generating attention maps. 

\subsection{Label-specific representation Learning}

Several methods use attention to extract semantic-specific features for interaction \cite{re8,re12,re25,re26} or contrastive learning \cite{re27}. Ye et al.\cite{re12} proposed an Attention‐Driven Dynamic GCN (ADD-GCN), in which a Semantic Attention Module (SAM) first extracts class‐specific features, and a Dynamic GCN (D-GCN) then models their relational structure. Zang et al.\cite{re25} generalized the single‐graph architecture into a multi‐graph framework based on ADD-GCN. Wu et al.\cite{re26} reformulated MLIC as a graph‐matching problem, they enhance SAM to identify critical regions and group them into instance $"$bags$"$, and perform instance–label matching via a graph‐matching scheme that integrates instance‐space and label‐semantic graphs. Dao et al.\cite{re27} utilized multi‐head attention over fused label embeddings to learn class‐specific representations, which are then projected into a latent space for contrastive learning. However, these methods often miss fine-grained details. To improve localization, Zhu et al.\cite{re28} designed a Class‐Specific Residual Attention (CSRA) module to better localize spatial regions corresponding to each class. Zhu et al.\cite{re13} further proposed a low‐rank bilinear model for object‐level representation, enhancing patch‐level representations through self‐attention over object cues to jointly leverage both local and global information. Still, their object localization remains coarse with irrelevant regions, mainly due to insufficient visual-semantic alignment.

\subsection{Visual-Semantic Alignment Methods}
Some studies adapt Transformer architectures from NLP to vision tasks\cite{re29,new_re30}. Alexey et al. \cite{re30} process image patches as tokens using multi-head self-attention, while Lanchantin et al. \cite{re14} model label-image interactions via C-Trans framework. These methods however incur high computational costs and rely heavily on precise token-label alignment. To overcome these limitations, Li et al. \cite{re34} reformulate multi-label classification as a conditional transport problem, aligning features through bidirectional transport cost minimization. Unlike traditional optimal transport requiring iterative Sinkhorn algorithm \cite{new_re33}, conditional transport uses semantic similarity with bidirectional planning, offering greater flexibility and lower complexity for deep learning integration. While advancing alignment, this approach fails to produce label-specific features.


\section{The Proposed Method}\label{sec3}
\subsection{Problem Definition and Overview}
Let $D = {(X_i, Y_i)}_{i=1}^B$ denote the training dataset of $B$ images, where $X_i \in \mathbb{R}^{H \times W \times 3}$ represents the $i$-th image and $Y_i = [y_i^1, \ldots, y_i^C] \in \{0,1\}^C$ is its multi-hot label vector over $C$ class labels. $y_i^j = 1$ indicates that the $i$-th image is annotated with the $j$-th class, otherwise, $y_i^j = 0$.



Figure 1 presents an overview of our proposed SCT framework, which consists of two primary components: \textbf{(1) Semantic-Related Feature Learning}, which integrates global image features with label embeddings to produce semantic-related features, and \textbf{(2) Feature Alignment}, which leverages bidirectional conditional transport. guided by a semantic map, to refine these features into discriminative label-specific representations. The final multi-label predictions are produced by aggregating region-level scores.
\begin{figure*}[H]  
    \centering
    \includegraphics[width=1\textwidth]{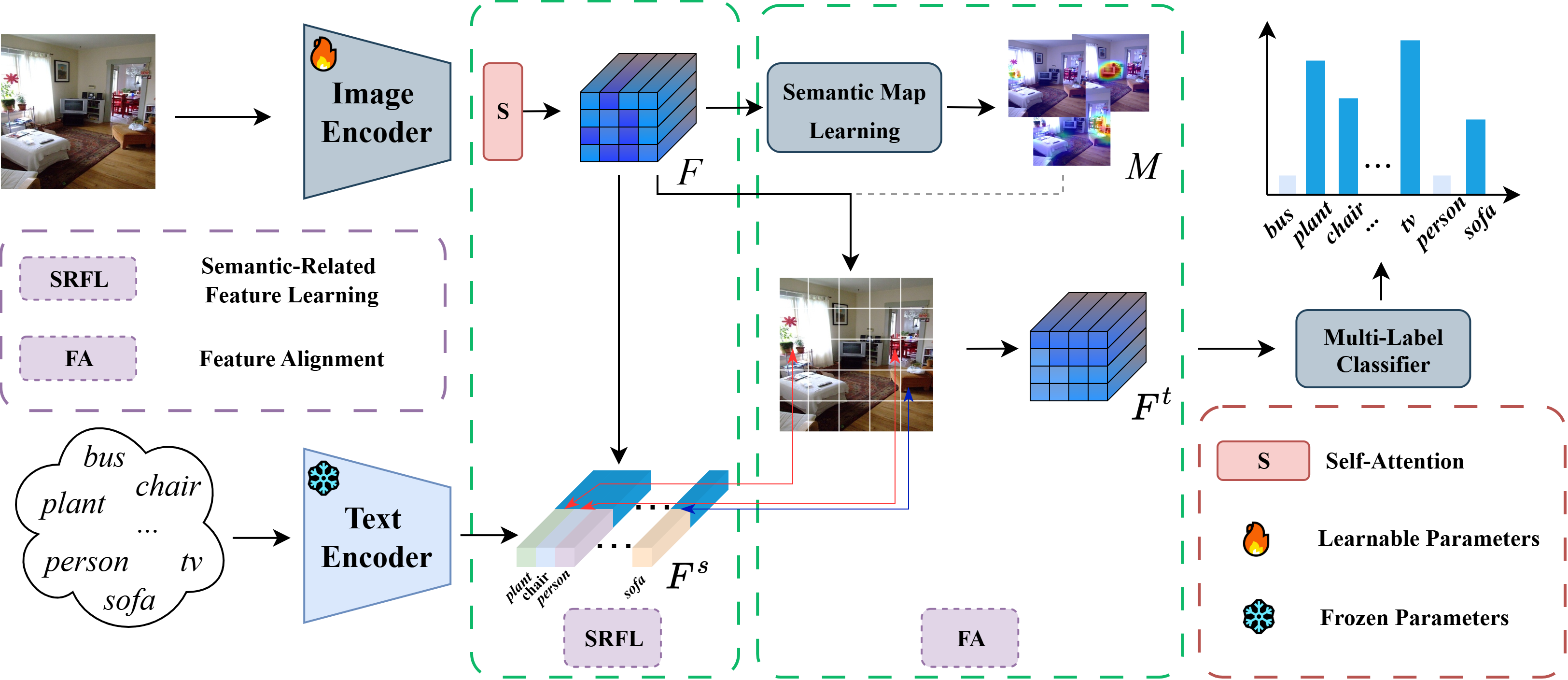}  
    \caption{Pipeline of our SCT Method.}
    \label{Fig.1}
\end{figure*}
\subsection{Semantic‐Related Feature Learning }
\textbf{Image features $F$ and Label embeddings $L$.} For a given image $X_i$, we employ a visual encoder (e.g., ResNet-101\cite{re35}) to produce an initial feature map. To overcome the constrained receptive fields of convolutional neural networks, we flatten this map and refine it with Transformer self-attention layers \cite{re29}, thereby yielding the image features $F_i\in R^{N \times d_v}$, where $N = H \times W$ denotes the total number of patches and $d_v$ is the feature channel dimension. For simplicity, the subscript $i$ is omitted hereafter. Concurrently, each class label $j$ is encoded into a label embedding vector $l_j \in R^{d_t}$ by a text encoder, forming the label embedding set $L=\{l_1, l_2, \ldots, l_C\}$.


\textbf{Semantic-Related Feature $F^S$.} Classifying directly from raw image features may lead the model to learn superficial statistical associations, failing to capture inherent label correlations and complex multi-label semantics. To address this limitation, we introduce a semantic-related feature learning module. Specifically, a global feature vector $F^G \in R^{d_v}$ is first obtained via global spatial pooling. It then serves as a feature-level label prompt to modulate the label embeddings $L$, yielding the semantic-related features $F^S \in R^{C \times d_v}$ as follows:
{
\setlength{\abovedisplayskip}{0.5pt}
\setlength{\belowdisplayskip}{0.5pt}
\begin{equation}
    F^{\mathrm{G}}=\mathrm{GSP}(F),
\end{equation}}
\vspace{-1.0\baselineskip} 
{
\setlength{\abovedisplayskip}{0.5pt}
\setlength{\belowdisplayskip}{0.5pt}
\begin{equation}
    F^{\mathrm{S}}=\text { Linear }\left(\operatorname{concat}\left(F^{\mathrm{G}}, L\right)\right),
\end{equation}}
where GSP($\cdot$) denotes global spatial pooling and Linear($\cdot$) represents a linear projection.

\subsection{Feature Alignment}
We transform the feature alignment as a conditional transport (CT) problem, learning an optimal transport matrix between the image features $F=\{f_{n}\}_{n=1}^{N}\in R^{N \times d_{v}}$ and the semantic-related features $F^{S}=\{f_{c}^{S}\}_{c=1}^{C}\in R^{C \times d_{v}}$. This is formalized by defining two discrete probability distributions over the shared feature space $X \in R^{d_v}$:
{
\setlength{\abovedisplayskip}{0.5pt}
\setlength{\belowdisplayskip}{0.5pt}
\begin{equation}
    \mathrm{P}=\sum_{n=1}^{N} \theta_{n} \delta_{f_{n}}, \mathrm{Q}=\sum_{c=1}^{C} \beta_{c} \delta_{f_{c}^{s}},
\end{equation}}
where $\mathrm{P}$ and $\mathrm{Q}$ denote the distributions of the image features and semantic-related features, respectively. Consider any points $f_n$ and $f_{c}^{s}$ in the shared space $X$, with $\delta_{f_{n}}$ and $\delta_{f_{c}^{s}}$ being the unit point masses at these coordinates. We define the discrete probability distributions using mass vectors $\theta \in \Sigma^{N}$ and $\beta \in \Sigma^C$, which lie in the probability simplices of $R^N$ and $R^C$, and must satisfy:
{
\setlength{\abovedisplayskip}{2pt}
\setlength{\belowdisplayskip}{2pt}
\begin{equation}
    \sum_{n=1}^{N} \theta_{n}=1,    
     \sum_{c=1}^{C} \beta_{c}=1,
\end{equation}}

The distance between the discrete distributions $\mathrm{P}$ and $\mathrm{Q}$ are quantified by the conditional transport (CT) distance, which is formulated as the following optimization problem:
\vspace{-1.0\baselineskip} 
{
\setlength{\abovedisplayskip}{2pt}
\setlength{\belowdisplayskip}{0.5pt}
\begin{equation}
    \mathrm{CT}(\mathrm{P}, \mathrm{Q})=\min _{\mathrm{T} \in \Pi(\theta, \beta)} \sum_{n, c} t_{n c} c o_{n c},
\end{equation}}
{
\setlength{\abovedisplayskip}{0.5pt}
\setlength{\belowdisplayskip}{02pt}
\begin{equation}
    \mathrm{Tl}{ }^{C}=\theta, \mathrm{T}^{\mathrm{T}} 1^{N}=\beta,
\end{equation}}
where $1^C$ and $1^N$ be all-ones vector, and 
$co_{nc} = \operatorname{cost}\bigl(f_{n}, f_{c}^{S}\bigr)$ 
represents the transport cost between $f_{n}$ and $f_{c}^{S}$. The conditional transport plan $\mathrm{T}$ is obtained by minimizing the CT cost.Inspired by \cite{re34}, we define the bidirectional conditional transport formulation:
{
\setlength{\abovedisplayskip}{0.5pt}
\setlength{\belowdisplayskip}{0.5pt}
\begin{equation}
    L_{\mathrm{CT}}(\mathrm{P}, \mathrm{Q})=\min _{\overrightarrow{\mathrm{T}} \overleftarrow{\mathrm{~T}}}\left(\sum_{n, c} 
    \overrightarrow{t}_{n c} c o_{n c}+\sum_{c, n} 
    \overleftarrow{t}_{c n} c o_{c n}\right),
\end{equation}}
where $\overrightarrow{t}_{nc}$ and $\overleftarrow{t}_{cn}$ denote the forward (from $f_n$ to $f_{c}^{S}$) and backward (from $f_{c}^{S}$ to $f_n$) conditional transport probabilities, respectively. The cost matrix $\mathrm{CO}$ is computed as follows:
{
\setlength{\abovedisplayskip}{0.5pt}
\setlength{\belowdisplayskip}{0.5pt}
\begin{equation}
   \mathrm{C O}=\textbf{1}-\frac{F \cdot\left(F^{\mathrm{S}}\right)^{\mathrm{T}}}{\|F\|\left\|F^{\mathrm{S}}\right\|},
\end{equation}}

For better spatial localization of object regions in the image features, we introduce a learnable semantic map $\mathrm{M} \in R^{N \times C}$, which is optimized by:
{
\setlength{\abovedisplayskip}{0.5pt}
\setlength{\belowdisplayskip}{0.5pt}
\begin{equation}
L_{\mathrm{m}}=L_{\operatorname{ASL}}\left(f_{\sigma}\left(\max_{j\in{\{1,2,\cdots,C\}}} \left(m_{\cdot j}\right)\right), Y\right),
\end{equation}}
where $L_{\mathrm{ASL}}{(\cdot,\cdot)}$ denotes ALS loss \cite{re36} and $\max\left(m_{\cdot j}\right)$ selects the highest activation score per class.


Therefore, the source distribution $\theta$ can be dynamically adjusted through the following transformation:
{
\setlength{\abovedisplayskip}{0.5pt}
\setlength{\belowdisplayskip}{0.5pt}
\begin{equation}
    \theta=\operatorname{softmax}\left(\mathrm{M} \frac{y}{\sum_{c} y^{c}}\right),
\end{equation}}
and the vector $\beta \in R^C$ in $\mathrm{Q}$ is obtained by normalizing the ground-truth label vector $y$:

\vspace{-1.0\baselineskip} 
{
\setlength{\abovedisplayskip}{0.5pt}
\setlength{\belowdisplayskip}{0.5pt}
\begin{equation}
    \beta=\operatorname{softmax}(y),
\end{equation}}

To achieve fine-grained alignment between visual and semantic features with high computational efficiency, we employ a low-rank bilinear model~\cite{re21} to compute the transport‐mass matrix $\mathrm{A}$, with elements $a_{n,c}$ representing the transport mass from the $n$-th image patch to the $c$-th class. The transport probability matrix $\mathrm{T}$ is then derived by scaling $\mathrm{A}$ with $\theta$ and $\beta$ under the constraints of Equation (6), as defined by: 
{
\setlength{\abovedisplayskip}{0.5pt}
\setlength{\belowdisplayskip}{0.5pt}
\begin{equation}
    \overrightarrow{t}_{n c}=\theta_{n} \frac{\exp \left(a_{n, c}\right)}{\sum_{c^{\prime}} \exp \left(a_{n, c^{\prime}}\right)},
\end{equation}}
{
\setlength{\abovedisplayskip}{0.5pt}
\setlength{\belowdisplayskip}{0.5pt}
\begin{equation}
    \overleftarrow{t}_{c n}=\beta_{c} \frac{\exp \left(a_{c, n}\right)}{\sum_{n^{\prime}} \exp \left(a_{c, n^{\prime}}\right)},
\end{equation}}

%

Finally, the semantic-aware representation \\
$F^{\mathrm{R}}=\left\{\left.f_{n}^{\mathrm{R}}\right|_{n=1} ^{N}\right\}\in R^{N \times d_{v}}$ is formulated as a transport-weighted feature fusion:
{
\setlength{\abovedisplayskip}{0pt}
\setlength{\belowdisplayskip}{2pt}
\begin{equation}
f_{n}^{\mathrm{R}}=\sum_{c=1}^{C} b_{n c} f_{c}^{\mathrm{S}},
\end{equation}}
where $f^\mathrm{S}_c$ denotes the $c$-th semantic‐related feature in $F^\mathrm{S}$ and $b_{n c}$ is computed as follows:
{
\setlength{\abovedisplayskip}{0.5pt}
\setlength{\belowdisplayskip}{0.5pt}
\begin{equation}
    b_{n c}=\frac{\exp \left(a_{n, c}\right)}{\sum_{c^{\prime}} \exp \left(a_{n, c^{\prime}}\right)},
\end{equation}}
\subsection{Training objectives}
For a given image, the multi-label prediction vector $z$ is computed by aggregating region-level scores: 
{
\setlength{\abovedisplayskip}{2pt}
\setlength{\belowdisplayskip}{2pt}
\begin{equation}
    \widetilde{w}_{n}=f_{\mathrm{mlc}}\left(f_{n}^{\mathrm{R}}\right),
\end{equation}}
\vspace{-1.0\baselineskip} 
{
\setlength{\abovedisplayskip}{2pt}
\setlength{\belowdisplayskip}{2pt}
\begin{equation}
    w_{n}=\operatorname{softmax}\left(\widetilde{w}_{n}\right),
\end{equation}}
\vspace{-1.0\baselineskip} 
{
\setlength{\abovedisplayskip}{2pt}
\setlength{\belowdisplayskip}{2pt}
\begin{equation}
    z=\sum_{n} w_{n} \cdot \widetilde{w}_{n},
\end{equation}}
where $f_\mathrm{mlc} (\cdot)$ is a multi‐label classifier, and the multi-label classification loss is defined as: 
{
\setlength{\abovedisplayskip}{0.5pt}
\setlength{\belowdisplayskip}{0.5pt}
\begin{equation}
    L_{\mathrm{cls}}=L_{\operatorname{ASL}}(z, y),
\end{equation}}
Finally, we obtain the total training loss by combining $L_{\mathrm{cls}}$, the semantic map confidence loss $L_{\mathrm{m}}$ and $L_{\mathrm{CT}}$: 
{
\setlength{\abovedisplayskip}{0.5pt}
\setlength{\belowdisplayskip}{0.5pt}
\begin{equation}
    L=L_{\mathrm{cls}}+\lambda_{1} L_{\mathrm{m}}+\lambda_{2} L_{\mathrm{CT}},
\end{equation}}
where $\lambda_1$ and $\lambda_2$ are hyperparameters. 

\section{Experiments}\label{sec4}
\subsection{Experimental Settings}
\textbf{Datasets.} To validate the effectiveness of our proposed SCT method, we utilize two widely used multi-label image datasets: PASCAL VOC 2007\cite{re37}(VOC2007) and MS-COCO\cite{re38}. Specifically, VOC2007 contains 20 categories, with an average of 1.5 labels per image, and consists of 5,011 training images and 4,952 test images. For MS-COCO, it includes 80 categories (with an average of 2.9 labels per image) and comprises 82,783 training images and 40,504 test images. 

\textbf{Evaluation criteria.} To evaluate the proposed method against baseline approaches, we adopt two standard metrics: Average Precision per class (AP) and mean Average Precision (mAP).
Additionally, we report the average overall precision (OP), recall (OR), F1 (OF1), as well as the average per-class precision (CP), recall (CR), F1 (CF1). For all metrics, higher values indicate better performance. 

\textbf{Experimental Details.} For fair comparison, we adopt ResNet‑101\cite{re35} pre‑trained on ImageNet as the backbone image encoder and BERT\cite{re39} pre‑trained on Wikipedia as the text encoder. We use the AdamW optimizer for training and apply RandAugment and Cutout\cite{re32} for data augmentation. To improve training stability, the Exponential Moving Average (EMA) with a decay rate of 0.9997 is employed. The self‑attention module contains a single layer with eight heads. For ASL loss, the hyperparameters $\gamma+$ and $\gamma-$ are set to 0 and 2, respectively. All experiments are performed on a single NVIDIA Tesla A40 GPU and our model is implemented using PyTorch. 

\subsection{Experimental Results Analysis}
\textbf{Results on VOC2007.} 
As shown in Table 1, which presents the performance comparison using the ResNet-101 backbone, our method outperforms all compared approaches, achieving 0.1$\%$, 0.7$\%$, and 0.8$\%$ higher mAP than SGRE-R101, C-TMS, and GM-MLIC respectively. Significant gains are observed in challenging categories like $'$table$'$ and $'$plant$'$. Compared to CLIP-based PatchCT\cite{re34} using MS-COCO pre-trained encoder with LoRA fine-tuning, our method reaches 97.3$\%$ mAP, surpassing PatchCT by 0.2$\%$. These results confirm our SCT method sets a new state-of-the-art in multi-label image recognition.

\begin{table*}
    \centering
    \footnotesize
   \caption{Experimental Results on VOC2007 in terms of class-wise average precision (AP, $\%$) and mean average precision (mAP, $\%$). Best scores are highlighted in \textbf{bold}. † indicates methods using high-resolution input (576×576) and * denotes methods using a CLIP image encoder.}
   \label{Table.1}

    \setlength{\tabcolsep}{1.4pt}       
    \renewcommand{\arraystretch}{1.0} 
    \begin{tabular}
    {@{\extracolsep\fill}|l|cccccccccccccccccccc|c|}
        \hline
        Method & plane & bike & bird & boat & bottle & bus & car & cat & chair & cow & table & dog & horse & moto & person & plant & sheep & sofa & train & tv & mAP \\
        \hline
        CNN-RNN~\cite{re18}       & 96.7 & 83.1 & 94.2 & 92.8 & 61.2 & 82.1 & 89.1 & 94.2 & 64.2 & 83.6 & 70.0 & 92.4 & 91.7 & 84.2 & 93.7 & 59.8 & 93.2 & 75.3 & 99.7 & 78.6 & 84.0 \\
        RLSD~\cite{re16}             & 96.4 & 92.7 & 93.8 & 94.1 & 71.2 & 92.5 & 94.2 & 95.7 & 74.3 & 90.0 & 74.2 & 95.4 & 96.2 & 92.1 & 97.9 & 66.9 & 93.5 & 73.7 & 97.5 & 87.6 & 88.5 \\
        HCP~\cite{re15}               & 98.6 & 97.1 & 98.0 & 95.6 & 75.3 & 94.7 & 95.8 & 97.3 & 73.1 & 90.2 & 80.0 & 97.3 & 96.1 & 94.9 & 96.3 & 78.3 & 94.7 & 76.2 & 97.9 & 91.5 & 90.9 \\
        RDAR~\cite{re19}             & 98.6 & 97.4 & 96.3 & 96.2 & 75.2 & 92.4 & 96.5 & 97.1 & 76.5 & 92.0 & 87.7 & 96.8 & 97.5 & 93.8 & 98.5 & 81.6 & 93.7 & 82.8 & 98.6 & 89.3 & 91.9 \\
        SSGRL$^{\dagger}$~\cite{re8} & 99.5 & 97.1 & 97.6 & 97.8 & 82.6 & 94.8 & 96.7 & 98.1 & 78.0 & 97.0 & 85.6 & 97.8 & 98.3 & 96.4 & 98.8 & 84.9 & 96.5 & 79.8 & 98.4 & 92.8 & 93.4 \\
        ML-GCN~\cite{re10}          & 99.5 & 98.5 & 98.6 & 98.1 & 80.8 & 94.6 & 97.2 & 98.2 & 82.3 & 95.7 & 86.4 & 98.2 & 98.4 & 96.7 & 99.0 & 84.7 & 96.7 & 84.3 & 98.9 & 93.7 & 94.0 \\
        TSGCN~\cite{re43}           & 98.9 & 98.5 & 96.8 & 97.3 & {87.5} & 94.2 & 97.4 & 97.7 & 84.1 & 92.6 & 89.3 & 98.4 & 98.0 & 96.1 & 98.7 & 84.9 & 96.6 & 87.2 & 98.4 & 93.7 & 94.3 \\
        ASL~\cite{re36}               & --   & --   & --   & --   & --   & --   & --   & --   & --   & --   & --   & --   & --   & --   & --   & --   & --   & --   & --   & --   & 94.4 \\
        CPCL~\cite{re44}             & 99.6 & 98.6 & 98.5 & 98.8 & 81.9 & 95.1 & 97.8 & 98.2 & 83.0 & 95.5 & 85.5 & 98.4 & 98.5 & 97.0 & 99.0 & 86.6 & 97.0 & 84.9 & 99.1 & 94.3 & 94.4 \\
        SST~\cite{re45}               & 99.8 & 98.6 & {98.9} & 98.4 & 85.5 & 94.7 & 97.9 & 98.6 & 83.0 & 96.8 & 85.7 & 98.8 & 98.9 & 95.7 & 99.1 & 85.4 & 96.2 & 84.3 & 99.1 & 95.0 & 94.5 \\
        CSRA~\cite{re28}             & {99.9} & 98.4 & 98.1 & 98.9 & 82.2 & 95.3 & 97.8 & 97.9 & 84.6 & 94.8 & 90.8 & 98.1 & 97.6 & 96.2 & 99.1 & 86.4 & 95.9 & 88.3 & 98.9 & 94.4 & 94.7 \\
        GM-MLIC~\cite{re46}       & 99.4 & 98.7 & 98.5 & 97.6 & 86.3 & {97.1} & 98.0 & {99.4} & 82.5 & 98.1 & 87.7 & {99.2} & 98.9 & {97.5} & {99.3} & 87.0 & {98.3} & 86.5 & 99.1 & 94.9 & 94.7 \\
        C-TMS~\cite{re33}            & 99.5 & 96.2 & 97.6 & 96.5 & 86.4 & 95.8 & 95.2 & 98.9 & {88.7} & 97.5 & 87.6 & {99.2} & {99.2} & 97.3 & 99.0 & 85.0 & 98.1 & 86.1 & 99.1 & 94.2 & 94.8 \\
        MCAR~\cite{re7}             & 99.7 & {99.0} & 98.5 & 98.2 & 85.4 & 96.9 & 97.4 & 98.9 & 83.7 & 95.5 & 88.8 & 99.1 & 98.2 & 95.1 & 99.1 & 84.8 & 97.1 & 87.8 & 98.3 & 94.8 & 94.8 \\
        SGRE-R101~\cite{re13}        & {99.9} & {99.0} & 98.4 & 98.8 & 82.2 & 96.7 & {98.3} & 98.8 & 85.0 & 97.3 & 89.7 & 98.7 & 98.7 & 97.3 & 99.1 & 86.9 & 97.7 & {89.6} & 99.2 & 95.9 & 95.4 \\
        
        \textbf{Ours-R101}           & {99.9} & 98.7 & 98.1 & 98.4 & 83.4 & 96.6 & 98.0 & 98.6 & 84.9 & 96.7 & {92.0} & 98.7 & 98.9 & 97.2 & 99.0 & {88.4} & 97.4 & 89.1 & 99.5 & {96.0} & {95.5} \\
        PatchCT*~\cite{re34}          & \textbf{100.0} & 99.4 & 98.8 & \textbf{99.3} & 87.2 & \textbf{98.6} & 98.8 & 99.2 & 87.2 & \textbf{99.0} & \textbf{95.5} & \textbf{99.4} & \textbf{99.7} & 98.9 & 99.1 & 91.8 & \textbf{99.5} & \textbf{94.5} & 99.5 & 96.3 & 97.1 \\
        \textbf{Ours*}                        & \textbf{100.0} & \textbf{99.6} & \textbf{98.9} & \textbf{99.3} & \textbf{89.9} & \textbf{98.6} & \textbf{99.2} & \textbf{99.5} & \textbf{90.1} & 98.4 & 94.2 & \textbf{99.4} & 99.2 & \textbf{99.0} & \textbf{99.7} & \textbf{93.0} & 99.3 & 93.2 & \textbf{99.6} & \textbf{96.5} & \textbf{97.3} \\ 
        \hline
    \end{tabular}
\end{table*}


\textbf{Results on MS-COCO.}  
Table 2 shows results with ResNet-101 backbone. Our method matches SGRE-R101 in mAP but performs better in other key measures like CP, OP, OR, and OF1. Against CLIP-based PatchCT\cite{re34} using the same setup, our method reaches 88.5$\%$ mAP, beating it by 0.2$\%$. These findings confirm our method's overall lead.



\begin{table*}
    \centering
    \footnotesize
    \caption{Experimental results on MS-COCO. Best scores are highlighted in \textbf{bold}. * denotes methods using a CLIP image encoder. All metrics in $\%$}
    \label{Table.2}

    \setlength{\tabcolsep}{6pt}       
    \renewcommand{\arraystretch}{1.0} 
    
    \begin{tabular}
    {@{\extracolsep\fill}|l|l|c|ccc|ccc|ccc|ccc|}
        \hline
        \multirow{2}{*}{Method} & \multirow{2}{*}{Resolution} & \multirow{2}{*}{mAP} & \multicolumn{3}{c|}{All} & \multicolumn{3}{c|}{All} & \multicolumn{3}{c|}{Top-3} & \multicolumn{3}{c|}{Top-3}\\
        & & & CP & CR & CF1 & OP & OR & OF1 & CP & CR & CF1 & OP & OR & OF1 \\ \hline
        CNN-RNN~\cite{re18}        & 224×224  & 61.2 & -    & -    & -    & -    & -    & -    & 66.0 & 55.6 & 60.4 & 69.2 & 66.4 & 67.8 \\
        ML-GCN~\cite{re10}         & 448×448  & 83.0 & 85.1 & 72.0 & 78.0 & 85.8 & 75.4 & 80.3 & 89.2 & 64.1 & 74.6 & 90.5 & 66.5 & 76.7 \\
        CMA~\cite{re24}            & 448×448  & 83.4 & 82.1 & 73.1 & 77.3 & 83.7 & 76.3 & 79.9 & 87.2 & 64.6 & 74.2 & 89.1 & 66.7 & 76.3 \\
        KSSNet~\cite{re49}         & 448×448  & 83.7 & 84.6 & 73.2 & 77.2 & 87.8 & 76.2 & 81.5 & --   & --   & --   & --   & --   & --   \\
        MCAR~\cite{re7}            & 448×448  & 83.8 & 85.0 & 72.1 & 78.0 & 88.0 & 73.9 & 80.3 & 88.1 & 65.5 & 75.1 & 91.0 & 66.3 & 76.7 \\
        TSGCN~\cite{re43}          & 448×448  & 83.5 & 81.5 & 72.3 & 76.7 & 84.9 & 75.3 & 79.8 & 84.1 & 67.1 & 74.6 & 89.5 & 69.3 & 69.3 \\
        GM-MLIC~\cite{re46}        & 448×448  & 84.3 & \textbf{87.3} & 70.8 & 78.3 & 88.6 & 74.8 & 80.6 & {90.6} & 67.3 & 74.9 & 94.0 & 69.8 & 77.8 \\
        CSRA~\cite{re28}           & 448×448  & 83.5 & 84.1 & 72.5 & 77.9 & 85.6 & 75.7 & 80.3 & 88.5 & 64.2 & 74.4 & 90.4 & 66.4 & 76.5 \\
        SST~\cite{re45}            & 448×448  & 84.2 & 86.1 & 72.1 & 78.5 & 87.2 & 75.4 & 80.8 & 89.8 & 64.1 & 74.8 & 91.5 & 66.4 & 76.9 \\
        TDRG~\cite{re31}           & 448×448  & 84.6 & 86.0 & 73.1 & 79.0 & 86.6 & 76.4 & 81.2 & 89.9 & 64.4 & 75.0 & 91.2 & 67.0 & 77.2 \\
        ASL~\cite{re36}            & 448×448  & 85.0 & --   & --   & {80.3} & --   & --   & {82.3} & --   & --   & --   & --   & --   & --   \\
        Q2L-R101~\cite{re32}       & 448×448  & 84.9 & 84.8 & 74.5 & 79.3 & 86.6 & 76.9 & 81.5 & 78.0 & {69.1} & 73.3 & 80.7 & 70.6 & 75.4 \\
        C-TMS~\cite{re33}          & 448×448  & 85.1 & 87.2 & 74.2 & 79.1 & 88.7 & 76.5 & 81.4 & 90.1 & 67.5 & 76.0 & 92.3 & {71.2} & {78.3} \\
        SGRE-R101~\cite{re13}      & 448×448  & {85.7} & 85.4 & {75.4} & 80.1 & 86.8 & 77.9 & 82.1 & 89.4 & 66.1 & 76.0 & 91.4 & 68.0 & 78.0 \\
        \textbf{Ours-R101}         & 448×448  & {85.7} & 85.7 & 75.2 & 80.1 & 86.9 & {78.0} & 82.2 & 89.9 & 66.1 & {76.2} & 91.5 & 68.1 & 78.1 \\
        PatchCT*~\cite{re34}        & 448×448  & 88.3 & 83.3 & 82.3 & \textbf{82.6} & 84.2 & 83.7 & \textbf{83.8} & \textbf{90.7} & \textbf{69.7} & \textbf{78.8} & 90.3 & 70.8 & \textbf{79.8} \\
        \textbf{Ours*}             & 448×448  & \textbf{88.5} & 80.4 & \textbf{83.7} & 82.0 & 81.0 & \textbf{86.4} & 83.6 & 87.4 & \textbf{69.7} & 77.6 & 89.8 & \textbf{71.5} & 79.6 \\ 
        \hline
    \end{tabular}
\end{table*}

\subsection{Ablation Study}
In this section, we conduct ablation studies on VOC2007 and MS-COCO to evaluate the effectiveness of each component in our proposed framework. For fair comparison, all experiments use ResNet-101 as the backbone network, and the detailed results are reported in Table 3. 

\begin{table}[htbp]
    \centering
    \footnotesize
    \setlength{\tabcolsep}{1.4pt}       
    \renewcommand{\arraystretch}{1.0} 
     \caption{Ablation study results of the components in our proposed method. The best-performing scores are highlighted in \textbf{bold}.}
    \label{Table.3}
    \begin{tabular}{lccc cccc c}
        \toprule
        \multirow{2}{*}{Method}               & \multicolumn{3}{c}{Components}      & \multicolumn{4}{c}{MS-COCO}            & VOC \\
        \cmidrule(lr){2-4} \cmidrule(lr){5-8} \cmidrule(lr){9-9}
                             & SA    & M and CT & GSP    & mAP   & CP    & CR    & CF1   & mAP  \\
        \midrule
        Baseline             &       &          &        & 83.5  & 84.7  & 71.6  & 77.6  & 94.5 \\
        Our w/ self-attn     & \checkmark &          &        & 85.4  & 86.0  & 74.5  & 79.8  & 95.0 \\
        Our w/ M + CT      & \checkmark & \checkmark &        & 85.6  & 86.8  & 74.3  & 80.1  & 95.2 \\
        Our SCT             & \checkmark & \checkmark & \checkmark & \textbf{85.7} & \textbf{85.7} & \textbf{75.2} & \textbf{80.1} & \textbf{95.5} \\
        \bottomrule
    \end{tabular}
\end{table}

We first test our baseline (ResNet-101 and classifier using region score aggregation), which performs comparably to TSGCN and CSRA, confirming our method works well. Adding a self-attention (SA) layer improves spatial context modeling. Next, we include both the semantic map and conditional transport alignment (M and CT). Finally, we add global spatial pooling (GSP) and concatenate its output with label embeddings.


Table 3 shows the self-attention module improves mAP, CP, CR and CF1 by 1.9$\%$, 1.3$\%$, 2.9$\%$ and 2.2$\%$ on MS-COCO, confirming its effectiveness in capturing spatial correlations. Adding semantic map and conditional transport (M and CT) further increases mAP by 0.2$\%$ while boosting precision and F1-score by 0.8$\%$ and 0.3$\%$. Finally, GSP brings additional gains of 0.1$\%$ on MS-COCO and 0.3$\%$ on VOC2007, demonstrating the combined benefits of feature-semantic alignment and global pooling.


Table 4 compares different pooling methods. Global max pooling (GMP) lowers mAP by 0.4$\%$ on both datasets as it keeps only the strongest responses. Global average pooling (GAP) works better by using all spatial information, proving more suitable for multi-label recognition.


\begin{table}[!ht]
    \centering
     \caption{Comparison results of global spatial pooling strategies in our proposed method}
    \setlength{\tabcolsep}{6pt}       
    \renewcommand{\arraystretch}{1.0} 
    \label{Table.4}
    \begin{tabular}{lccccc}
        \toprule
        \multirow{2}{*}{Method} & \multicolumn{4}{c}{MS-COCO} & VOC \\
        \cmidrule(lr){2-5} \cmidrule(lr){6-6}
               & mAP  & CP   & CR   & CF1  & mAP  \\
        \midrule
        GAP    & 85.7 & 85.7 & 75.2 & 80.1 & 95.5 \\
        GMP    & 85.3 & 87.2 & 73.6 & 79.8 & 95.1 \\
        \bottomrule
    \end{tabular}
\end{table}
\subsection{Visualization Analysis}

This section presents the visualization analysis performed on VOC2007 using OpenCV. To evaluate the capability of the proposed model in localizing discriminative object regions, we visualize the label-specific attention maps generated by our SCT method and SGRE, as shown in Figure 2. 


Figure 2 shows SGRE's attention maps, while generally focusing on relevant areas, contain noticeable noise, for example incorrectly highlighting lamps and pillows when identifying a sofa. In contrast, our SCT method precisely pinpoints key components like dog and motorbike.Figure 3 demonstrates how our semantic map M first identifies roughly relevant regions, which the conditional transport mechanism then refines into exact object locations when recognizing 'motorbike' and 'person'.

\begin{figure}[t]  
    \centering
    \includegraphics[width=0.5\textwidth]{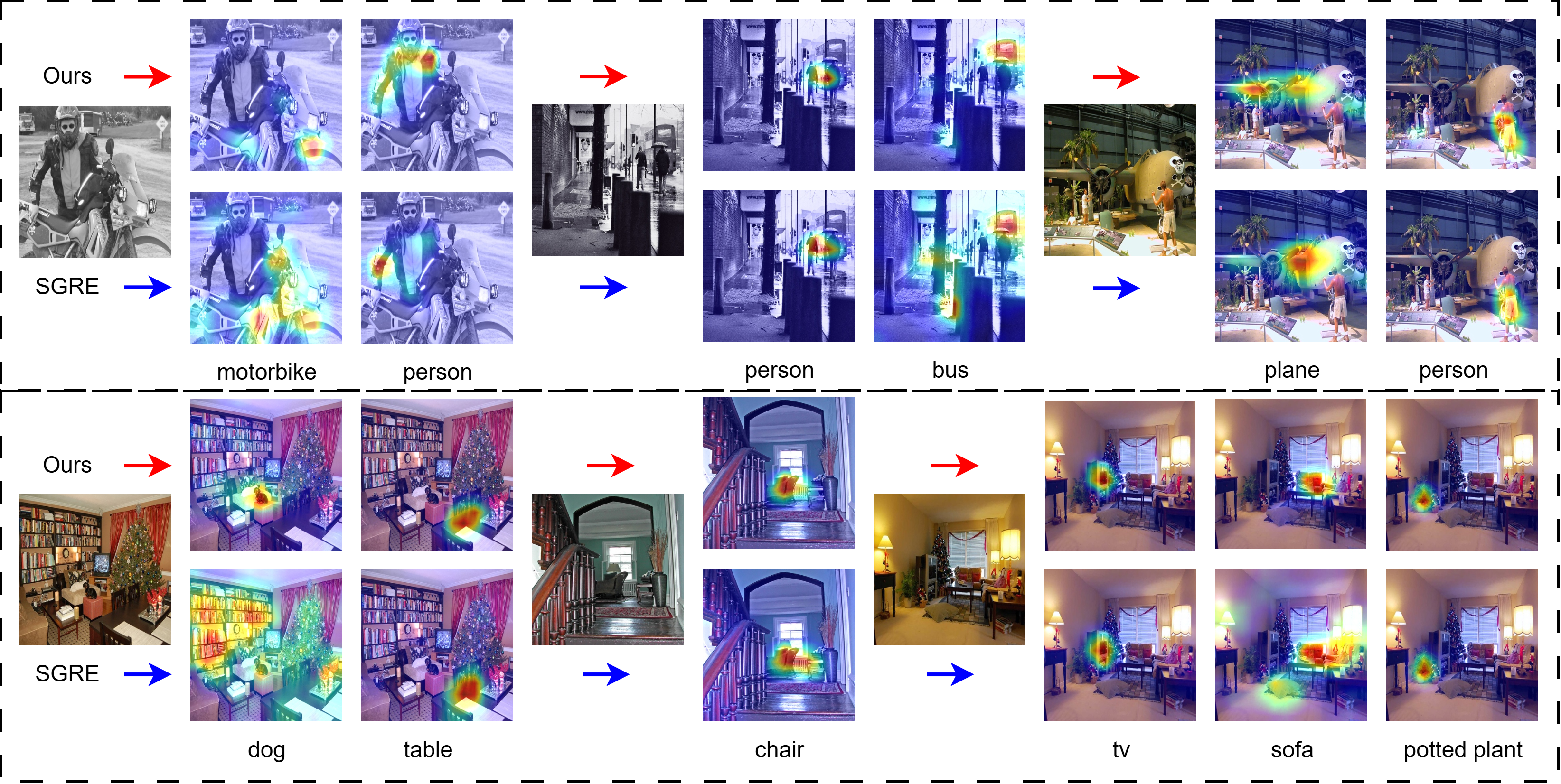}  
    \caption{Visual analysis of the SGRE method and our SCT method.}
    \label{Fig.2}
\end{figure}


Moreover, SCT demonstrates the ability to identify annotation errors in the dataset. As shown in Figure 4, SCT correctly identifies labels missing from ground-truth annotations, particularly for challenging cases involving low-visibility, occlusion, or high difficulty in recognition. This further validates the robustness and effectiveness of our proposed SCT framework.


\begin{figure}[t]  
    \centering
    \includegraphics[width=0.5\textwidth]{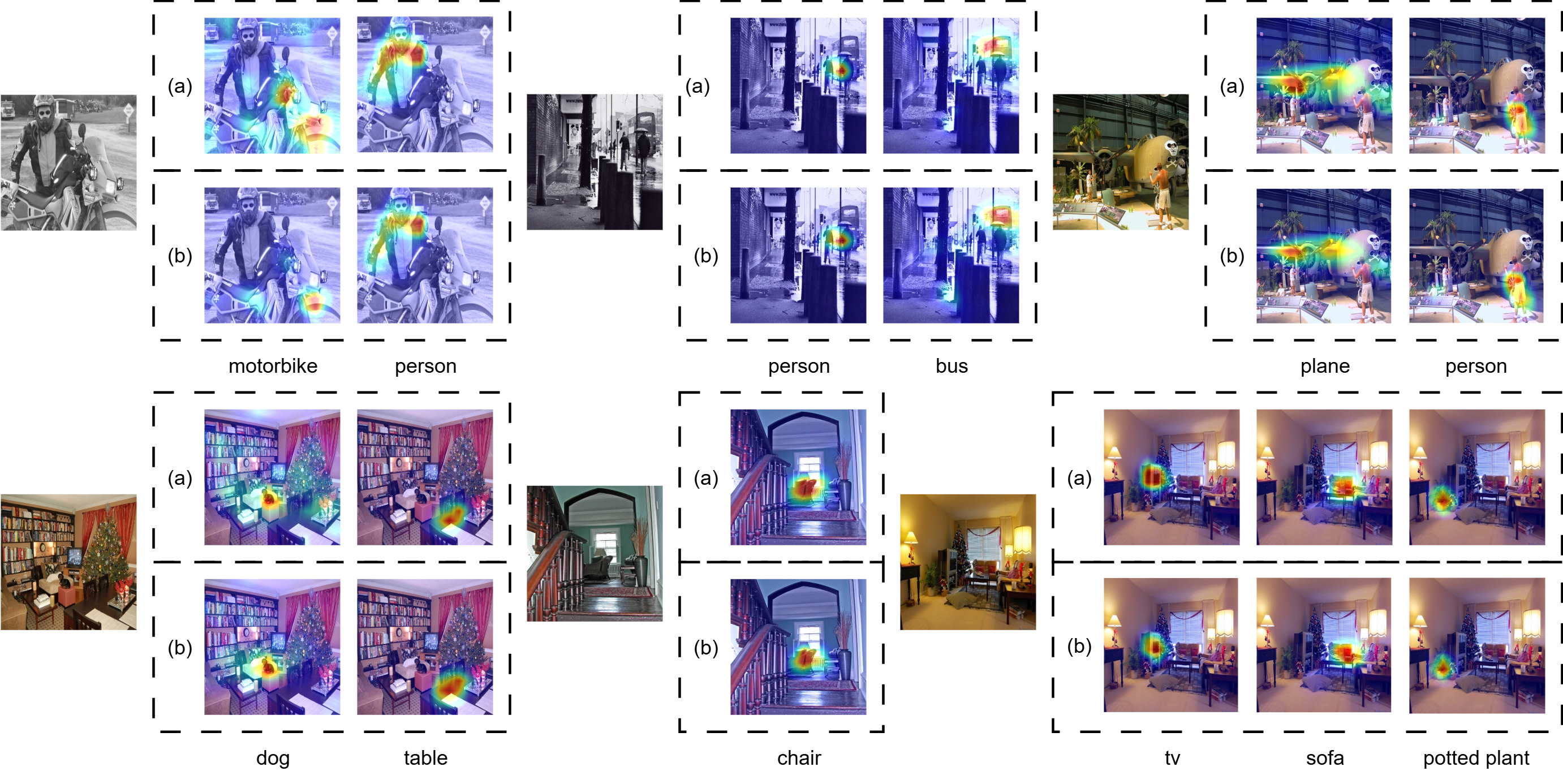}  
    \caption{Visual analysis of our SCT method. (a) Class activation maps from semantic map $\mathrm{M}$; (b) Class activation maps from semantic-aware representation $F^t$.}
    \label{Fig.3}
\end{figure}

\section{Conclusions}\label{sec5}

This paper proposes a novel semantic-aware representation learning via conditional transport method for multi-label image classification. By integrating semantic-aware representation learning with a conditional transport-based alignment mechanism, our approach effectively localizes discriminative object regions while capturing complex semantic relationships among labels. The framework implicitly models label correlations through the fusion of global image features and label embeddings, thereby obtaining highly discriminative label-specific feature representations. Experimental results on benchmark multi-label image datasets validate the effectiveness of our proposed method. In future work, this framework will be further extended to multi-label few-shot and zero-shot learning scenarios.

\begin{figure}[t]  
    \centering
    \includegraphics[width=0.4\textwidth]{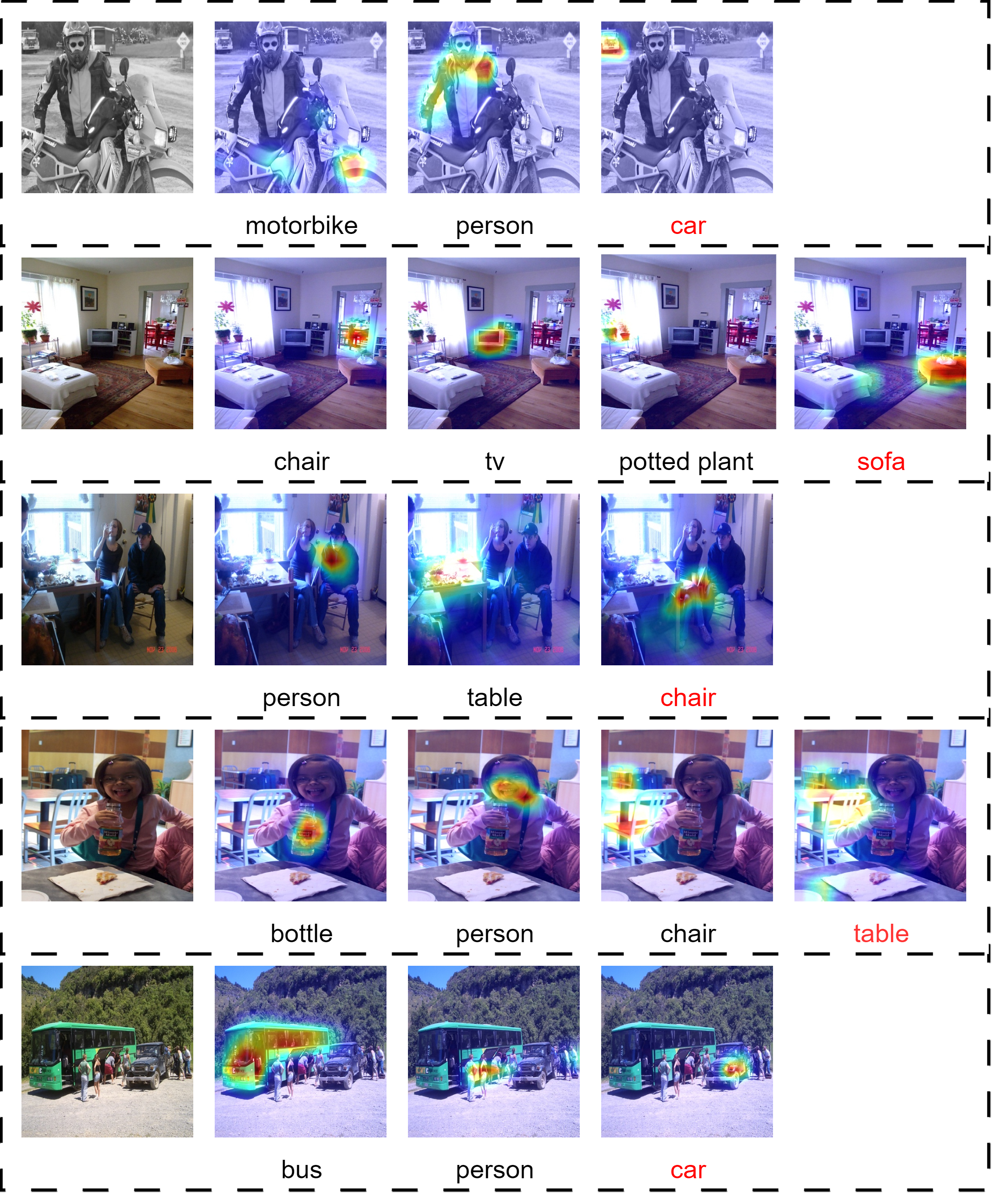}  
    \caption{Visualization of labels recognized by SCT. Labels in black font denote ground truth annotations from the dataset, while labels in red font indicate additional labels identified by the SCT method.}
    \label{Fig.4}
\end{figure}

\section*{Acknowledgements}
This work is supported by National Natural Science Foundation of China under Grants 62041604, 62172198, 61762064, 62063029, Natural Science Foundation of Jiangxi Province under Grant 20232BAB202047, and Scientific Startup Foundation for Doctors under Grant EA202107235. The authors would like to thank the anonymous referees and the editors for their helpful comments and suggestions.







\end{document}